\documentclass[letterpaper]{article} 
\usepackage{aaai23}  
\usepackage{times}  
\usepackage{helvet}  
\usepackage{courier}  
\usepackage[hyphens]{url}  
\usepackage{graphicx} 
\usepackage{natbib}  
\usepackage{caption} 
\usepackage{multirow}
\usepackage{amsmath,amssymb,amsfonts}
\usepackage{algorithm}
\usepackage{algorithmic}

\usepackage{newfloat}
\usepackage{listings}
\DeclareCaptionStyle{ruled}{labelfont=normalfont,labelsep=colon,strut=off} 
\floatstyle{ruled}
\newfloat{listing}{tb}{lst}{}
\floatname{listing}{Listing}
\title{Fine-grained Retrieval Prompt Tuning}

\author{
    Shijie Wang \textsuperscript{\rm 1},
    Jianlong Chang \textsuperscript{\rm 2},
    Zhihui Wang \textsuperscript{\rm 1},
    Haojie Li \textsuperscript{\rm 1,3}\thanks{Corresponding author: hjli@dlut.edu.cn.},
    Wanli Ouyang \textsuperscript{\rm 4},
    Qi Tian \textsuperscript{\rm 2}
}
\affiliations{
    \textsuperscript{\rm 1}International School of Information Science \& Engineering, Dalian University of Technology, China \\
    \textsuperscript{\rm 2} Huawei Cloud \& AI, China \\
    \textsuperscript{\rm 3}College of Computer and Engineering, Shandong University of Science and Technology, China \\
    \textsuperscript{\rm 4}SenseTime Computer Vision Research Group, The University of Sydney, Australia
    
}

\usepackage{bibentry}

\begin{document}

\maketitle

\begin{abstract}
Fine-grained object retrieval aims to learn discriminative representation to retrieve visually similar objects. 
However, existing top-performing works usually impose pairwise similarities on the semantic embedding spaces or design a localization sub-network to continually fine-tune the entire model in limited data scenarios, thus resulting in convergence to suboptimal solutions.
In this paper, we develop Fine-grained Retrieval Prompt Tuning (FRPT), which steers a \textit{frozen} pre-trained model to perform the fine-grained retrieval task from the perspectives of sample prompting and feature adaptation. 
Specifically, FRPT only needs to learn \textit{fewer parameters} in the prompt and adaptation instead of fine-tuning the entire model, thus solving the issue of convergence to suboptimal solutions caused by fine-tuning the entire model.
Technically, a discriminative perturbation prompt (DPP) is introduced and deemed as a sample prompting process, which amplifies and even exaggerates some discriminative elements contributing to category prediction via a content-aware inhomogeneous sampling operation.
In this way, DPP can make the fine-grained retrieval task aided by the perturbation prompts close to the solved task during the original pre-training. Thereby, it preserves the generalization and discrimination of representation extracted from input samples.
Besides, a category-specific awareness head is proposed and regarded as feature adaptation, which removes the species discrepancies in features extracted by the pre-trained model using category-guided instance normalization. And thus, it makes the optimized features only include the discrepancies among subcategories.
Extensive experiments demonstrate that our FRPT with fewer learnable parameters achieves the state-of-the-art performance on three widely-used fine-grained datasets.

\end{abstract}
\section{Introduction}
Fine-grained Object Retrieval (FGOR) is to retrieve images belonging to various subcategories of a certain meta-category (\textit{i.e.}, birds, cars and aircraft) and return images with the same subcategory as the query image.
However, retrieving visually similar objects is still challenging in practical applications, especially when there exists large intra-class variances but small inter-class differences. 
As a result, the key to FGOR lies in learning the discriminative and generalizable embeddings to identify the visually similar objects.

Recently, the successful FGOR works fight against large intra-class but small inter-class variances by designing specialized metric constraints \cite{DBLP:conf/eccv/TehDT20,DBLP:conf/cvpr/WangHHDS19,DBLP:conf/eccv/BoudiafRZGPPA20} or locating object and even parts \cite{DBLP:journals/tip/WeiLWZ17,DBLP:conf/ijcai/ZhengJSWHY18,DBLP:conf/wacv/MoskvyakMDB21}. Although metric-based and localization-based works can learn discriminative embeddings to identify fine-grained objects, the FGOR model learned from last phase is still needed to be fine-tuned on next phase endlessly, forcing the model to adapt the fine-grained retrieval task. However, continually fine-tuning the FGOR model could result in convergence to suboptimal solutions especially when facing limited-data regimes, inevitably limiting the retrieval performance \cite{DBLP:conf/cvpr/HuangZGS21, DBLP:conf/icml/ZintgrafSKHW19}. Therefore, a question naturally arises: is it possible that we can still learn discriminative embeddings without fine-tuning the entire FGOR model? It is already answered yes by natural language processing (NLP) with prompt techniques.

Prompt-based learning \cite{DBLP:journals/corr/abs-2208-10159} is a task-relevant instruction prepended to the input to adapt the downstream tasks to the frozen pre-trained models.
Its key idea is to reformulate the downstream tasks aided by an appropriate prompt design, making it close to those solved during the original pre-training.
Thereby, prompt-based learning could utilize pre-trained models directly instead of fine-tuning them to adapt downstream tasks.
Following this ideology, the vision-language pre-training task has been gradually developed, which diametrically obtains visual guidance concepts from natural language via putting visual category semantics into text inputs as prompts \cite{DBLP:conf/icml/JiaYXCPPLSLD21, DBLP:conf/iccv/KamathSLSMC21, DBLP:conf/icml/RadfordKHRGASAM21}.
Though these works achieve remarkable performance on many downstream vision tasks, their prompt tuning strategies are tailored for the multi-modal model and thus inapplicable to fine-grained vision models. Therefore, how to devise a prompt scheme for fine-grained vision models to solve the issue  of convergence to suboptimal solutions caused by fine-tuning the entire FGOR model is worthy of investigation. 

To this end, we propose Fine-grained Retrieval Prompt Tuning (FRPT), which equips with  discriminative perturbation prompt (DPP), pre-trained backbone model, and category-specific awareness head (CAH). 
FRPT merely learns fewer parameters in DPP and CAH while freezing the weights of backbone model, thus solving the convergence to suboptimal solutions. 
Specifically, as a sample prompting process, DPP is designed to zoom and even exaggerate some elements contributing to category prediction via a content-aware inhomogeneous sampling operation.
In this way, DPP can adjust the object content towards facilitating category prediction, which makes the FGOR task prompted with this discrimiantive content perturbation close to the solved task during the original pre-training.
Nevertheless, a non-negligible problem is that the backbone model without fine-tuning will focus on extracting features to answer the question,``{\it what are the different characteristics between species}" instead of ``{\it how to distinguish fine-grained objects within the same meta-category}". 
Therefore, CAH is regarded as feature adaptation to optimize the features extracted by the backbone model via removing the species discrepancies using category-guided instance normalization, thus making the optimized features only contain the discrepancies among subcategories.
Unlike fine-tuning, FRPT has fewer parameters to train, but still learns embeddings with greater discrimination and generalization owing to DPP and CAH, thus solving the convergence to suboptimal solutions caused by fine-tuning the entire model.

Our main contributions are summarized as below:
\begin{itemize}
    \item  We propose FRPT to steer a frozen pre-trained model to perform FGOR task from the perspectives of sample prompting and feature adaptation. To the best of our knowledge, we are the first to develop the prompt tuning scheme specifically for handling the convergence to suboptimal solutions caused by fine-tuning strategy in FGOR.\looseness=-1

 \item A discriminative perturbation prompt is proposed to emphasize the elements contributing to decision boundary, which instructs the frozen pre-trained model to capture subtle yet discriminative details.

\item A category-specific awareness head is designed to remove the discrepancies among species, which makes the features specifically for identifying fine-grained objects within the same meta-category.

\item FRPT only needs to optimize about 10\% of the parameters than being fully fine-tuned, and even achieves the new state-of-the-art results, {\it e.g.}, 3.5\% average Recall@1 increase on three widely-used fine-grained retrieval datasets. 
\end{itemize}


\section{Related Work}
\textbf{Prompting Tuning.}
Prompting \cite{DBLP:journals/corr/abs-2210-00788, DBLP:journals/corr/abs-2207-14381,DBLP:journals/corr/abs-2208-10159} in NLP reformulates the downstream dataset into a language modeling problem, so that a frozen language model directly adapts to a new task. Therefore, prompt tuning has now been applied to handle a variety of NLP tasks, including language understanding and generation \cite{DBLP:conf/emnlp/LesterAC21, DBLP:conf/acl/LiL20, DBLP:journals/corr/abs-2110-07602, DBLP:journals/tacl/JiangXAN20}.
Recently, prompt tuning has been introduced into multi-modal computer vision \cite{ DBLP:conf/icml/RadfordKHRGASAM21,
DBLP:journals/corr/abs-2210-15929, DBLP:journals/corr/abs-2109-11797}. CPT \cite{DBLP:journals/corr/abs-2109-11797} converts visual grounding as the problem of filling in the blank by creating visual prompts with colored blocks and color-based textual prompts.
However, these prompt tuning based multi-modal works focus on extending the capabilities of a language-based model, inapplicable to pre-trained vision models. To fill this gap, we are the first work to develop a parameter-efficient FRPT via introducing the sample prompting and feature adaptation, thus instructing the frozen pre-trained vision model to perform FGOR task.

\textbf{Fine-grained Object Retrieval:} Existing FGOR methods can be roughly divided into two groups. The first group, \textit{localization-based schemes} focuses on localizing the objects or even parts from images via exploring the activation of features \cite{DBLP:journals/tip/WeiLWZ17,DBLP:conf/ijcai/ZhengJSWHY18, DBLP:journals/mms/WangWWWL22, DBLP:conf/wacv/MoskvyakMDB21,DBLP:conf/aaai/WangWLO22}. CRL \cite{DBLP:conf/ijcai/ZhengJSWHY18} designs an attractive object feature extraction strategy to facilitate the retrieval task. Despite the inspiring achievement, the shortcoming of these works is that they only focus on discriminative embeddings while neglect the inter-class and intra-class correlations between subcategories, thus reducing the retrieval performance. Therefore, the second group, \textit{metric-based schemes} is learning an embedding space where similar examples are attracted, and dissimilar examples are repelled \cite{DBLP:conf/eccv/TehDT20,DBLP:conf/cvpr/WangHHDS19,DBLP:conf/eccv/BoudiafRZGPPA20, DBLP:conf/iccv/KoGK21, DBLP:conf/icml/RothMOCG21, DBLP:conf/iccv/ZhengZL021}. 
However, these existing approaches continually fine-tune the entire representation model in limited-data regimes, resulting in convergence to suboptimal solutions. To cope with this issue, FRPT attaches fewer learnable parameters into the frozen backbone network to train, instead of fine-tuning the entire representation model, but still learns discriminative embeddings.

\section{Fine-grained Retrieval Prompt Tuning}
 
\begin{figure*}[t]
\begin{center}
   \includegraphics[width=1\linewidth]{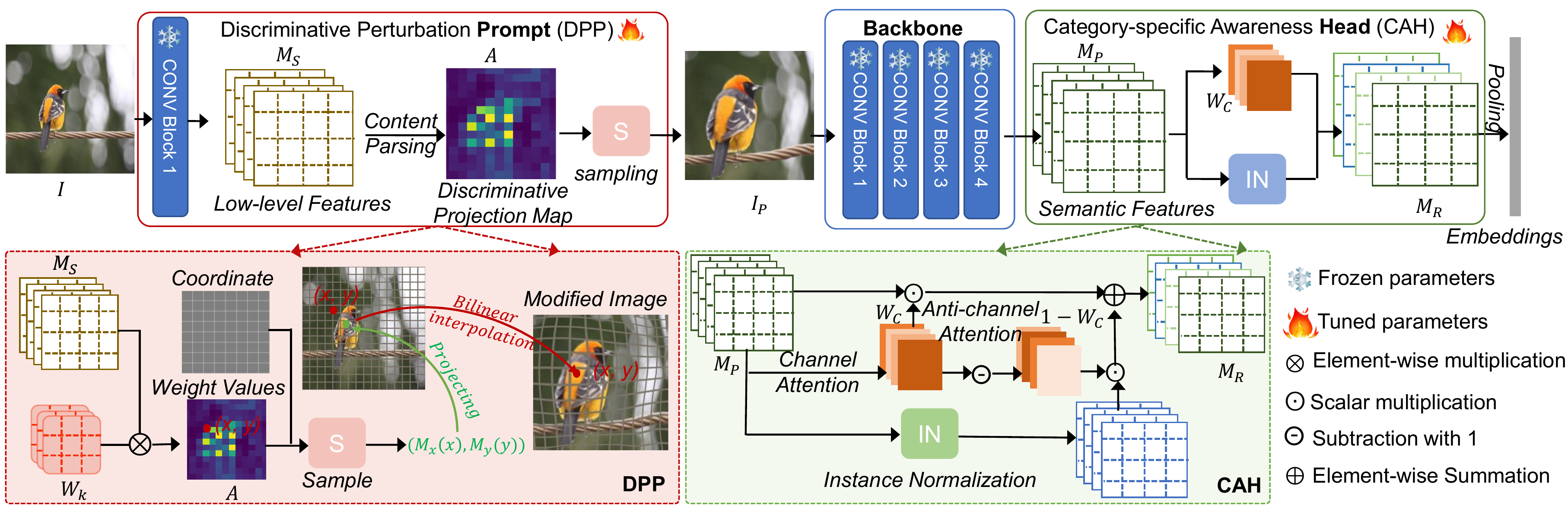}
\end{center}
   \caption{FRPT consists of three essential modules: the discriminative perturbation prompt module to zoom and even exaggerate the discriminative elements within objects, the frozen pre-trained backbone network to extract rich object representation, and the category-specific awareness head to compress the species discrepancies within the extracted semantic features.}

\end{figure*}


 
We propose Fine-grained Retrieval Prompt Tuning (FRPT) for steering frozen pre-trained model to perform FGOR task. FRPT only optimizes fewer learnable parameters within sample prompting and feature adaption, and keeps the backbone frozen during training. In this way, FRPT solves the issue of convergence to suboptimal solutions caused by fine-tuning the entire FGOR model.

\subsection{Network Architecture}
The network architecture is given in Fig. 1.
Formally, given an input image $ I $, we feed it into the discriminative perturbation prompt (DPP) module to generate the modified image $ I_P $ which selectively highlights certain elements contributing to decision boundary. After that, we take $ I_P $ as input to the frozen pre-trained backbone, and thus the semantic features $ M_P $ are outputted. 
To make $M_P$ identify fine-grained objects within the same meta-category rather than recognizing diverse species, we feed $ M_P $ into the category-specific awareness head (CAH) module to generate the category-specific features $ M_R $.  Finally, the category-specific features are pass through a global average pooling operation to obtain the discriminative embeddings and then apply them to search other samples with the same subcategory. 


\subsection{Discriminative Perturbation Prompt}
To solve the issue of convergence to suboptimal solutions caused by fine-tuning the entire model, we are inspired by prompt-based learning to solely modify the pixels in inputs, which makes the fine-grained retrieval task with the prompts close to those solved during pre-training.
Therefore, we propose a discriminative perturbation prompt (DPP) module to zoom and even exaggerate some elements contributing to category prediction in the pixel space. In this way, DPP can steer the frozen pre-trained model with this perturbed prompt to perceive much more discriminative details, thus bringing the high-quality representation.
Concretely, DPP consists of two steps. The first step, \textit{content parsing}, is to learn a discriminative projection map which reflects the location and intensity of discriminative information, and the second step, \textit{discriminative content modification}, is that zooms discriminative elements via performing the content-aware inhomogeneous sampling operation on each input image under the guidance of the discriminative projection map. 

\textbf{Content parsing.} Perceiving details and semantics plays a vital role in perturbing the object content \cite{DBLP:conf/mm/WangWLO20,DBLP:conf/aaai/WangLWO21, DBLP:conf/icmcs/WangWZLL19}. Upon this, we design a content parsing module to perceive the locations and scales of discriminative semantics and details from the low-level features. The content parsing has an appealing property: large field of view that can aggregate contextual information within a large receptive field instead of exploiting pixel neighborhood. Thereby, the content parsing can capture the discriminative semantics from the low-level details while preserving the discriminative details.

Given an input image $ I \in \mathbb{R}^{3\times H \times W} $, we  feed $ I $ into the convolutional block 1 of frozen pre-trained representation model $ \mathcal{F}_{block1} $ to generate the low-level features $ M_S \in\mathbb{R}^{C_S \times H_S \times W_S} $, where $ H_S $, $W_S $, $ C_S $ are the height, width and number of channels. It should be clarified that since the shallow layers in pre-trained representation model are sensitive to low-level details, such as color and texture, their parameters are not required to be updated and still work well.


After obtaining the low-level features $ M_S $, we transform them into the discriminative projection map in a content-aware manner. Concretely, each target location on the discriminative projection map $ \mathcal{A} \in \mathbb{R}^{H_S \times W_S} $ corresponds to $ \sigma^2 $ source locations on $ M_S $. Therefore, each target location shares a content-aware kernel $ W_k \in \mathbb{R}^{ \sigma \times \sigma \times C_S } $, where $ \sigma $ is the content-aware kernel size and is not less than the $ \frac{1}{2} $ width of $ M_S $, and thus is set to 31 in our experiment settings. With the shared content-aware kernel $ W_k $, the content parsing module will specify the locations, scales and intensities of discriminative semantics and details. For a target location $ (m, n) $, the calculation formulation is shown in Eqn. (1), where $ r = \lfloor \sigma/2\rfloor $: 
\begin{equation}
\mathcal{A}_{(m,n)} = \sum_{w=-r}^r\sum_{h=-r}^r \sum_{c=1}^{C_S} W_k^{(w,h,c)} \cdot M_S^{(m+w, n+h,c)}.
\end{equation}

Before being applied to the discriminative content modification operation, the discriminative projection map $ \mathcal{A} $ is normalized with a softmax function spatially. The normalization step forces the sum of the weight values in $ \mathcal{A} $ to 1:
\begin{equation}
 \mathcal{A}_{ij}= \frac{e^{\mathcal{A}_{ij}}}{\sum_{i=1}^{W_S}\sum_{j=1}^{H_S}e^{\mathcal{A}_{ij}}}.
\end{equation} 

\textbf{Discriminative content modification.}
This module utilizes the spatial information of sample points and corresponding sample weights in the discriminative projection map to rearrange object content, which further highlights some elements contributing to category prediction in inputs. Therefore, the modified image $ I_P \in\mathbb{R}^{3\times W \times H} $ can be formulated as below:
\begin{equation}
I_P = \mathcal{S}(I, \mathcal{A}),
\end{equation}
where $ \mathcal{S}(\cdot) $ indicates the content-aware inhomogeneous sampling function. 

Our basic idea for inhomogeneous sampling is considering the discriminative projection map $ \mathcal{A} $ as probability mass function, where the area with large sample weight value in $ \mathcal{A} $ is more likely to be sampled. Therefore, we compute a mapping function between the modified and original images and then use the grid sampler introduced in STN \cite{DBLP:conf/nips/JaderbergSZK15} to rearrange objects. The mapping function can be decomposed into two dimensions, i.e., horizontal and vertical axis dimensions, thus reducing mapping complexity. Taking the coordinate $ (x, y) $ in the modified image for example, we can calculate the mapping coordinate $(\mathcal{M}_x(x), \mathcal{M}_y(y))$ in the original input as below:
\begin{equation}
\begin{aligned}
&\mathcal{M}_x(x) =  \\
& \frac{\sum_{w = 1 }^{W_S}\sum_{h = 1 }^{H_S}\mathcal{A}(w, h)\cdot \mathcal{D}<(\frac{x}{W_S},\frac{y}{H_S}),(\frac{w}{W_S},\frac{h}{H_S})> \cdot \frac{w}{W_S}}{\sum_{w = 1 }^{W_S}\sum_{h = 1 }^{H_S}\mathcal{A}(w, h)\cdot \mathcal{D}<(\frac{x}{W_S},\frac{y}{H_S}),(\frac{w}{W_S},\frac{h}{H_S})>},
\end{aligned}
\end{equation}
\begin{equation}
\begin{aligned}
&\mathcal{M}_y(y) = \\ &\frac{\sum_{w = 1 }^{W_S}\sum_{h = 1 }^{H_S}\mathcal{A}(w, h)\cdot \mathcal{D}<(\frac{x}{W_S},\frac{y}{H_S}),(\frac{w}{W_S},\frac{h}{H_S})> \cdot \frac{h}{H_S}}{\sum_{w = 1 }^{W_S}\sum_{h = 1 }^{H_S}\mathcal{A}(w, h)\cdot \mathcal{D}<(\frac{x}{W_S},\frac{y}{H_S}),(\frac{w}{W_s},\frac{h}{H_S})>},
\end{aligned}
\end{equation}
where $ \mathcal{D}<,> $ is a Gaussian distance kernel to act as a regularizer to avoid some extreme cases, such as all pixels converge to the same location. According to Eqn. (4)(5), we can find that that each spatial location of the modified image requires a global perspective to select the filled pixel in the original input and thus reserve the structure knowledge. In addition, the regions with large sample weight values are allocated with more sampling chances, thus zooming and even exaggerating the discriminative elements in inputs. More importantly, each pixel in the modified image is correlated with each other, and the object structure is slightly perturbed instead of being completely destroyed.

After obtaining the mapping coordinates, we then use the differentiable bi-linear sampling mechanism proposed in STN, which linearly interpolates the values of the 4-neighbors (top-left, top-right, bottom-left, bottom-right) of $ (\mathcal{M}_x(x), \mathcal{M}_y(y)) $ to approximate the final output, denoted by $ I_P $,
\begin{equation}
I_P(x,y) = \sum_{(i,j)\in\mathcal{N}(\mathcal{M}_x(x), \mathcal{M}_y(y))} w_p \cdot I(i,j),
\end{equation}
where $ \mathcal{N}(\mathcal{M}_x(x), \mathcal{M}_y(y)) $ denotes neighbors of the mapping point $ \mathcal{M}_x(x), \mathcal{M}_y(y) $ in $ I $, and $ w_p $ is the bi-linear kernel weights estimated by the distance between the mapping point and its neighbors.

\subsection{Category-specific Awareness Head}
The modified image $ I_P $ is fed into the frozen pre-trained representation model to extract the semantic features $ M_P \in \mathbb{R}^{C_P\times H_P \times W_P} $ derived from its last convolution layer, where $ H_P, W_P, C_P $ are the height, width, and dimension of the semantic features.
However, since the semantic features extracted by the frozen pre-trained representation model aim to distinguish diverse species instead of fine-grained objects within the same meta-category, they still include some discrepancies between species, thus reducing the generalization ability of representation \cite{DBLP:conf/cvpr/WangWYLLL20,  DBLP:conf/mm/WangWZLZL19, DBLP:conf/aaai/WangWLDL20}. 
To handle this issue, we design a category-specific awareness head (CAH) to remove the species discrepancies as much as possible while only keeping the features most relevant to subcategories. 

The core of CAH is Instance Normalization (IN) \cite{DBLP:conf/eccv/PanLST18} guided by the subcategory supervision signals to remove the discrepancy among species. However, directly utilizing IN may damage discriminative information, inevitably affecting object retrieval performance. To deal with this limitation, we design a channel attention-guided IN to select the features containing the species discrepancies based on the channel-wise attention, remove them using IN, and integrate the original discriminative and optimized features into category-specific features $M_R$:
\begin{equation}
M_R = W_C \cdot M_P + (1-W_C) \cdot IN(M_P),
\end{equation}
where $ W_C \in \mathbb{R}^{C_P} $ denotes the weight coefficients that indicates the importance of diverse channel features, and $ IN(M_P) $ is the instance-normalized features of input $ M_P $.
Inspired by SENet \cite{DBLP:conf/cvpr/HuSS18}, the channel-wise attention can be provided by 
\begin{equation}
W_C = \Delta(W_L \delta(W_F g(M_P))),
\end{equation}
where $ g(\cdot) $ represents the global average pooling operation, $ W_F \in \mathbb{R}^{\frac{C_P}{r}\times C_P}$ and $ W_L \in \mathbb{R} ^{C_P \times \frac{C_P}{r}} $ are learnable parameters in the two bias-free fully-connected layers which are followed by ReLU activation function $ \delta $ and a sigmoid activation function $ \Delta $. For the dimension reduction ratio $ r $, we aim to balance the performance and complexity and thus set it to 8.
The parameter-free IN is defined as 
\begin{equation}
IN(M_P^i) = \frac{M_P^i - E[M_P^i]}{\sqrt{Var[M_P^i]+\epsilon}},
\end{equation}
where $ M_P^i \in \mathbb{R}^{H_P \times W_P}$ is the $ i $-th channel of feature map $ M_P $, $ \epsilon $ is used to avoid dividing-by-zero, the mean $ E[\cdot] $ and standard-deviation $ Var[\cdot] $ are calculated per-channel.

It should be clarified that since the channel-wise attention in CAH is guided using the category signal $y$, it would determine the relevant features about the discrepancies among subcategories and the irrelevant visual patterns which do not contribute to the prediction of $y$. We argue that the irrelevant features do not only contain the useless discrepancies among species but also some implicitly vital information contributing to subcategory prediction. Therefore, we use IN to remove these discrepancies among species and preserve the left vital information instead of directly discarding the irrelevant features, further obtaining much better performance.

\subsection{Optimization}

After obtaining the category-specific features, we train the model with the cross-entropy loss merely. The following cross-entropy loss is imposed on classifier $ C(\cdot) $ to predict the subcategories:
\begin{equation}
\mathcal{L} = - log P(y|C(g(M_R)|\theta)),
\end{equation} 
where $ y $ denotes the label, and $ C(g(M_R)|\theta) $ is the predictions of the classifier with parameters $ \theta $. The optimization process only affects the parameters within the DPP and CAH modules, but leaves no impact on backbone network during backward propagation, thus attacking the issue concerning with the convergence to suboptimal solutions owing to fine-tuning the entire representation model.

\section{Experiments}
\noindent\textbf{Datasets.} CUB-200-2011 \cite{Branson2014Bird} contains 200 bird subcategories with 11,788 images. We utilize the first 100 classes (5,864 images) in training and the rest (5,924 images) in testing. The Stanford Cars \cite{Krause20133D} contains 196 car models of 16,185 images. The spilt in Stanford Cars \cite{Krause20133D} is also similar to CUB, which is split into the first 98 classes (8,045 images) for training and the remaining classes (8,131 images) for testing. FGVC Aircraft \cite{DBLP:journals/corr/MajiRKBV13} is divided into first 50 classes (5,000 images) for training and the rest 50 classes (5,000 images) for testing.

\textbf{Implementation Details.}
We apply the widely-used Resnet \cite{He2015Deep} in our experiments with the pre-trained parameters. The input raw images are resized to $ 256 \times 256 $ and cropped into $ 224 \times 224 $. We train our models using Stochastic Gradient Descent (SGD) optimizer with weight decay of 0.0001, momentum of 0.9, and batch size of 32. 
We adopt the commonly used data augmentation techniques, \textit{i.e.}, random cropping and erasing, left-right flipping, and color jittering for robust feature representations. The total number of training epochs is set to 500. Our model is relatively lightweight and is trained end-to-end on two NVIDIA 2080Ti GPUs for acceleration.
The initial learning rate is set to $ 10^{-3} $, with exponential decay of 0.9 after every 50 epochs. 

\textbf{Evaluation protocols.} We evaluate the retrieval performance by \textit{Recall@K} with cosine distance, which is average recall scores over all query images in the test set and strictly follows the setting in \citet{DBLP:conf/cvpr/SongXJS16}. Specifically, for each query, our model returns the top $ K $ similar images. In the top $ K $ returning images, the score will be 1 if there exists at least one positive image, and 0 otherwise.

\subsection{Ablation Study}

\begin{table}\centering
	
	\begin{tabular}{c|c|c|c}
		\hline
		\hline
		Method & Params &CUB & Cars \\
		\hline
		Pre-training (PT) & 0M & 44.9\% & 58.9\% \\
		Fine-tuning &  23.53M & 69.5\% & 84.4\% \\
		\hline
		DPP + PT &  0.25M  & 55.7\%& 72.7\% \\
		PT + CAH &  2.62M & 61.8\% & 76.9\% \\
		DPP + PT + CAH(w/o IN) &  2.86M & 72.2\% & 89.6\%\\
		DPP + PT + CAH &  2.86M & 74.3\% & 91.1\%\\
		\hline
		\hline
		
	\end{tabular}\\
	\caption{ The ablative retrieval results (Recall@1) of different variants of our method. We test the models on CUB-200-2011 (CUB) and Stanford Cars (Cars). IN denotes Instance Normalization. Params denotes the numbers of learnable   parameters. 
	}
\end{table}

We conduct some ablation experiments to illustrate the effectiveness of the proposed modules. The baseline method uses ResNet-50 as the backbone network, followed by an Fully-Connection (FC) layer as the classifier and trained with the cross-entropy loss $ \mathcal{L} $ in the same setting.
As shown in Tab. 1, the contribution of each component is revealed. We first verify the performance with freezing and fine-tuning parameters of backbone network, respectively. The results can reflect directly freezing backbone network significantly reduces the retrieval performance. Compared with pre-trained model, DPP only introduces few learnable parameters and improves the \textit{Recall@1} accuracy by 10.8\% and 15.8\% on CUB-200-2011 and Stanford Cars datasets. The improvements also prove that DPP can instruct pre-trained models to perform fine-grained retrieval task by perturbing the object content. Additionally, we add the CAH module into the pre-trained model, bringing in 16.7\% and 18.0\% performance gains on two datasets. By this means that CAH could remove the species discrepancies from the features derived from pre-trained vision model, further improving retrieval results. When these two modules work together to learn discriminative embeddings, the \textit{Recall@1} accuracy is significantly improved by 29.4\% and 32.2\%. Additionally, the improving performance also verifies the importance of optimizing the features containing the species discrepancies via IN rather than directly discarding them. Compared with the fine-tuning strategy, our prompt scheme has only 2.86 million learnable parameters and even suppresses fine-tuning with performance gains of 4.6\% and 6.7\% on two datasets, respectively.
These results demonstrate that the proposed DPP and CAH can steer the frozen pre-trained model to perfrom the fine-grained retrieval task.

\subsection{Comparison with the State-of-the-Art Methods}

\begin{table*}\centering
	
	\begin{tabular}{c|c||cccc|cccc|cccc}
		\hline
		\hline
       \multirow{2.5}{*}{Method} & \multirow{2.5}{*}{Arch}& \multicolumn{4}{|c|}{CUB-200-2011}&  \multicolumn{4}{c|}{Stanford Cars 196} & \multicolumn{4}{c}{FGVC Aircraft}\\
		\cline{3-14}
		&  & 1&2&4&8& 1&2&4&8& 1&2&4&8 \\
		\hline
		\hline
		SCDA (\citeauthor{DBLP:journals/tip/WeiLWZ17})& R50& 57.3 & 70.2 & 81.0 &88.4& 48.3& 60.2&71.8&81.8& 56.5& 67.7&77.6&85.7\\
PDDM (\citeauthor{DBLP:journals/tog/Bala15})& R50& 58.3&69.2&79.0&88.4&57.4&68.6&80.1&89.4&-&-&-&-\\

CRL (\citeauthor{DBLP:conf/ijcai/ZhengJSWHY18}) & R50& 62.5 & 74.2 & 82.9 &89.7& 57.8& 69.1&78.6&86.6& 61.1& 71.6&80.9&88.2\\
HDCL (\citeauthor{DBLP:journals/ijon/ZengLWZCL21}) & R50& 69.5 & 79.6 & 86.8 &92.4& 84.4& 90.1&94.1&96.5& 71.1& 81.0&88.3&93.3\\
\hline 
\hline
		DGCRL (\citeauthor{DBLP:conf/aaai/ZhengJSZWH19})& R50& 67.9 & 79.1 & 86.2 &91.8& 75.9& 83.9&89.7&94.0& 70.1& 79.6&88.0&93.0\\
		DCML (\citeauthor{DBLP:conf/cvpr/ZhengWL021}) & R50& 68.4 & 77.9 & 86.1 &91.7& 85.2& 91.8&96.0&98.0&-& -& -&-\\
		DRML (\citeauthor{DBLP:conf/iccv/ZhengZL021}) & In3 & 68.7 & 78.6 & 86.3 & 91.6 & 86.9 & 92.1& 95.2& 97.4 & -& -& -&-\\
		CEP (\citeauthor{DBLP:conf/eccv/BoudiafRZGPPA20})& R50&69.2 & 79.2 & 86.9 &91.6& 89.3& 93.9&96.6&98.1&  -& -& -&-\\
		MemVir (\citeauthor{DBLP:conf/iccv/KoGK21}) & R50 & 69.8&-&-&-& 86.4&-&-&-&- &- &- &-\\
	    IBC (\citeauthor{DBLP:conf/icml/SeidenschwarzEL21}) & R50 & 70.3 & 80.3& 87.6& 92.7& 88.1& 93.3& 96.2& 98.2& - &- &- &- \\
		S2SD (\citeauthor{DBLP:conf/icml/RothMOCG21})& R50& 70.1 & 79.7& -&-&89.5& 93.9&-&-&- &- &- &-\\
		ETLR (\citeauthor{DBLP:conf/cvpr/KimKCK21})& In3& 72.1 & 81.3 & 87.6 &-& 89.6& 94.0&96.5&-&-& -& -&- \\
		PNCA++ (\citeauthor{DBLP:conf/eccv/TehDT20})& R50& 72.2 & 82.0 & 89.2 &93.5& 90.1& 94.5&97.0&98.4&-& -& -&-\\
\hline
\hline

Our FRPT & R50& \textbf{74.3} & \textbf{83.7} & \textbf{89.8} &\textbf{94.3}& \textbf{91.1}& \textbf{95.1}&\textbf{97.3}&\textbf{98.6}& \textbf{77.6}& \textbf{85.7}&\textbf{91.4}&\textbf{95.6}\\
\hline
\hline
	\end{tabular}
	\caption{ Comparison of different methods on CUB-200-2011, Stanford Cars 196 and FGVC Aircraft datasets. "Arch" denotes  the architecture of using backbone network. "R50" and "In3" represent Resnet50 \cite{He2015Deep} and Inception V3 \cite{DBLP:conf/cvpr/SzegedyVISW16}, respectively. 
} 
\end{table*}

We compare our FRPT with state-of-the-art (SOTA) fine-grained object retrieval approaches. In Tab. 2, the performance of different methods on CUB-200-2011, Stanford Cars-196, and FGVC Aircraft datasets is reported, respectively. In the table from top to bottom, the methods are separated into three groups, which are (1) localization-based networks, (2) metric-based frameworks, and (3) our FRPT.

Existing works tend to localize object or parts to directly improve the discriminative ability of representation. Despite the encouraging achievement, the existing works still have limited ability in learning discriminative features across different subcategories due to only paying more attention to the optimization of discriminative features while overlooking the inter-class and intra-class correlations between subcategories. Therefore, the success behind these models based on deep metric learning can be largely attributed to being able to precisely identify the negative/positive pairs via enlarging/shrinking their distances, which indirectly explores the discriminative ability of features. 
However, these works learn discriminative embeddings via continually fine-tuning the entire representation model in limited-data regimes. We argue that this could result in convergence to suboptimal solutions and inevitably reducing model's generalization capacity. To attack this issue, we propose FRPT to attach fewer learnable parameters of the sample prompting and feature adaptation into the frozen pre-trained model to perform the FGOR task.
Therefore, FRPT can achieve significantly better retrieval performance than existing works using the fine-tuning strategy.


\subsection{Discussions}

\begin{table}\centering

	\begin{tabular}{c|c|c||c|c|c}
		\hline
       \hline
       Method & Arch & Params & CUB & Cars & Air \\

		\hline
		\hline
		Fine-tuning & R50 & 23.5M & 69.5 & 84.2 & 70.1 \\
		Our FRPT  & R50 & 2.9M &74.3 & 91.1& 77.6\\
		Fine-tuning & R101 & 43.6M & 70.9 & 85.1 & 69.7 \\
		Our FRPT  & R101 & 2.9M &75.6 & 90.4& 76.2\\
		\hline
		\hline

\end{tabular}
	\caption{ Comparison of fine-tuning strategy on CUB-200-2011 (CUB),  Stanford Cars 196 (Cars) and FGVC Aircraft (Air) datasets.  "R50" and "R101" represent Resnet50 and Resnet101\cite{He2015Deep}, respectively. 
} 
\end{table}

\textbf{Effect of prompt tuning strategy.} As listed in Tab. 3, fine-tuning the pre-trained vision models can degrade retrieval performance compared to freezing them. This phenomenon is reasonable since fine-tuning the pre-trained models on limited fine-grained datasets may hurt their capability of generally visual modelling due to convergence to suboptimal solutions. 
Additionally, our FRPT enjoys the consistent improvements on three object retrieval datasets, which validates stronger generalization ability of our sample adaptation prompts and feature adaptation head. Besides, when the pre-trained vision model is switched from Resnet50 to Resnet101, our FRPT does not introduce more learnable parameters and takes full advantage of the stronger representation power of larger models, thus resulting in the improvement of retrieval performance again.
To better display the positive impact of our FRPT, we visualize the retrieval accuracy and training loss curves in Fig. 2. As can be observed from our FRPT curves, the increasing number of training epochs generally brings slow performance improvement and significantly increases the convergence speeds. One important reason of this phenomenon is that our FRPT only introduces fewer learnable parameters and thus attacks the issue concerning with convergence to the suboptimal solutions.

\textbf{Effective few-shot learning.} To deeper explore the effectiveness of FRPT, we conduct extensive experiments based on the few-shot setting with two different numbers of samples per subcategory: 10 and 5 on CUB-200-2011. Across the 5-shot and 10-shot experimental setting in Tab. 4, our FRPT consistently outperforms the fine-tuning strategy with different pre-trained vision models. Compared with fine-tuning pre-trained models using all images in CUB-200-2011, our FRPT only use 10 samples per subcategory but obtain the proximity performances. Since our FRPT only needs to learn a few parameters and attack the issue concerning with convergence to suboptimal solutions accordingly, our method further achieves better results than the fully fine-tuning strategy when facing only a few training samples. Therefore, the above results demonstrate the outperforming performance of FRPT owning to attaching few but effective parameters into the frozen backbone network.

\begin{table}\centering

	\begin{tabular}{c|c|c||c|c}
		\hline
		\hline
		Method & Arch & Params & 10-shot & 5-shot \\
		\hline
		\hline
		Fine-tuning & R50 &  23.5M  & 63.1\% & 59.7\%  \\ 
		Our FRPT & R50 &  2.9M  & 66.6\% &  62.8\%  \\
		Fine-tuning & R101 & 43.6M  & 64.7\%  & 61.5\% \\ 
		Our FRPT & R101 & 2.9M & 68.9\% & 65.2\% \\
		\hline
		\hline

\end{tabular}
	\caption{  Recall@1 results on CUB-200-2011 about few-shot learning. 10-shot and 5-shot indicate that only 10 and 5 images per category are used during training, respectively.
} 
\end{table}

\begin{figure*}[t]
\begin{center}
  \includegraphics[width=0.93\linewidth]{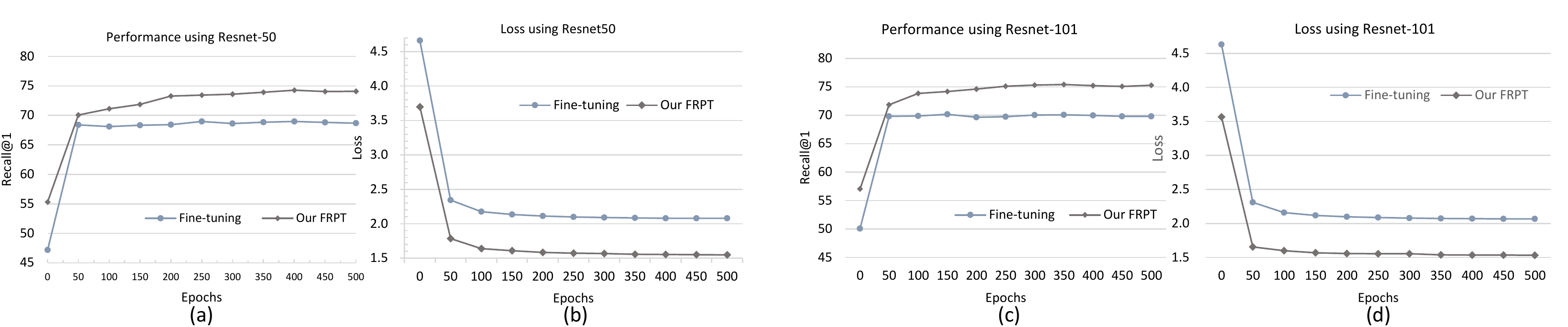}
\end{center}
  \caption{Curve visualization based on CUB-200-2011. (a) (c) denote the recall@1 curves about retrieval performance using Resnet-50 and Resnet-101, respectively. (b)(d) are the loss curves using Resnet-50 and Resnet-101, respectively.
  }

\end{figure*}

\begin{table}\centering

\begin{tabular}{c||cccc}
\hline \hline
Method & R@1 & R@2 & R@4 & R@8 \\
\hline 
CAM & 63.7\% & 74.3\% & 82.5\%  &89.7\% \\
Bounding box & 67.6\% & 79.3\%& 85.8\%& 91.6\% \\
\hline
Our DPP &74.3\% & 83.7\% & 89.8\% &94.3\% \\
\hline \hline
\end{tabular}
\caption{Performance comparison with other prompts in terms of Recall@K on CUB-200-2011.}
\end{table}

\begin{figure}[!t]
\centering

    \includegraphics[width = 0.9\linewidth]{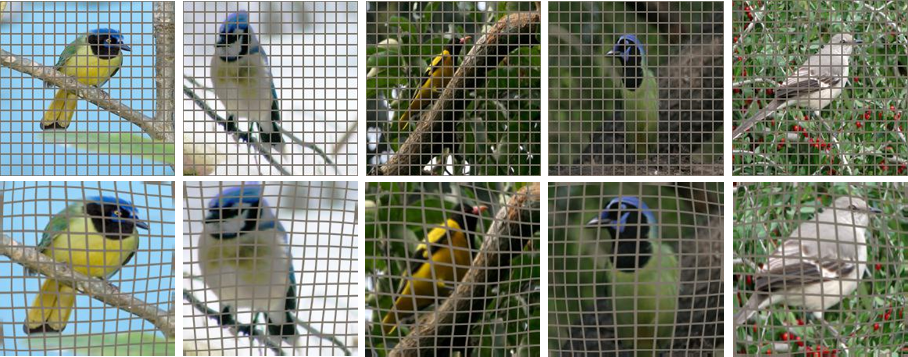}
    \caption{Visualization of the object content perturbation. The first and second rows denote the original and modified images, respectively.}
    
\end{figure}
\begin{figure}[!t]
\centering
    \includegraphics[width = 0.9\linewidth]{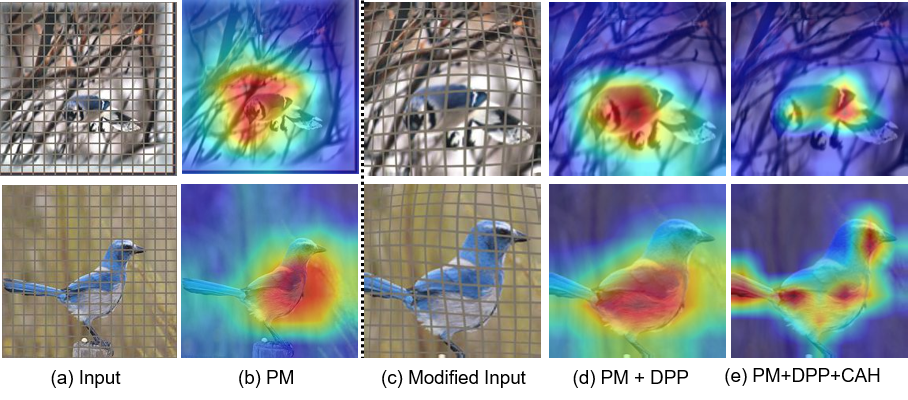}
    \caption{ Illustration of class activation maps (CAM). (a)(c) are the original and modified images, respectively. (b)(d)(e) denote CAMs generated by diverse networks.}

\end{figure}

\textbf{Fixed prompts vs. Learnable prompts.} More insight into the prompt scheme can be obtained by simple switching the processing manner of input images. As can be seen from Tab. 5, switching the processing method from the discriminative perturbation prompt (DPP) to the fixed prompt strategy, \textit{i.e.} directly zooming objects, leads to a significant performance drop.
Concretely, we use the class activation map (CAM) or the bounding boxes provided by the annotation information to localize the objects and then crop them from the original images. Tab. 5 shows a significant performance improvement when we utilize more accurate localization manner to remove the background information and preserve more object regions as much as possible. However, our FRPT zooms and even exaggerates the idiosyncratic elements contributing to decision boundary rather than to simply amplify objects and remove background, thus making the FGOR task aided by the discriminative perturbation prompt close to the solved task during the original pre-training and forming a steady improvement.

\textbf{What makes a network retrieve objects visually?} With this question in our mind, we exhibit the visualization results of original and modified images in Fig. 3. These visualization images can interpret why and how our approach can correctly identify diverse subcategories. As shown in the second row, our sample prompting scheme can enhance the visual evidence of object parts via the dense sampling operation while suppressing the background and even non-discriminative parts, thus instructing the pre-trained model to pay more attention to discriminative details and improving the retrieval performance accordingly. 
It should be clarified that we manually put grid lines on the images to better display the pixel shift in the images after our prompt processing. In Fig. 4, in addition to showing the original and modified images, we present the discriminative activation maps of three representation models, i.e, pre-trained model (Fig. 4(b)) , our FRPT without CAH (Fig. 4(d)), and our FRPT (Fig. 4(e)). It is clear that using DPP module can make the network focus on the object rather than background information, thus improving the discriminative ability of feature representation. Compared to Fig. 4(d), the activation maps (e) can pay more attention to the category-specific details via introducing CAH module. Based on these visualizations, our model generates clearer object boundaries and emphasises the discriminative details, thus providing higher retrieve performance.

\section{Conclusion}
In this paper, we propose Fine-grained Retrieval Prompt Tuning (FRPT), which aims to solve the issue of convergence to sub-optimal solutions caused by fine-tuning the entire FGOR model. FRPT design the discriminative perturbation prompt (DPP) and category-specific awareness head (CAH) to steer frozen pre-trained vision model to perform fine-grained retrieval task. 
Technically, DPP zooms and exaggerates some pixels contributing to category prediction, which assists the frozen pre-trained model prompted with this content perturbation to focus on discriminative details. 
CAH optimizes the semantic features extracted by pre-trained model via removing the species discrepancies using category-guided instance normalization,  which makes the optimized features sensitive to fine-grained objects within the same meta-category.
Extensive experiments demonstrate that our FRPT with fewer learnable parameters achieves the state-of-the-art performance on three widely-used fine-grained datasets.
\section*{Acknowledgements}
This work is supported in part by the National Natural Science Foundation of China under Grant NO. 61976038 and NO.61932020.
\bibliography{anonymous-submission-latex-2023}
\end{document}